\documentclass[10pt]{article}
\usepackage[preprint]{tmlr}
\usepackage[T1]{fontenc}
\usepackage[utf8]{inputenc}
\usepackage{amsmath,amssymb}
\usepackage{booktabs}
\usepackage{microtype}
\usepackage{graphicx}
\usepackage{pgfplots}
\pgfplotsset{compat=1.18}

\usepackage{hyperref}
\usepackage{url}

\title{Equal-Budget Monte Carlo Tree Search versus Randomized-Restart List Scheduling for DAG Workflows: A Controlled Equal-Budget Computational Study}

\author{\name Yuyu Zhang\thanks{Unfunded personal research, conducted independently on personal time and equipment, unrelated to any employment duty. Correspondence: 771262402@qq.com}}

\begin{document}
\maketitle

\begin{abstract}
Monte Carlo Tree Search (MCTS) is often proposed for scheduling directed-acyclic-graph (DAG) workflows, yet most studies never run the comparison that could refute it---giving a much simpler method the same search effort. We run that comparison under a controlled protocol. All arms receive an identical budget---the same number of candidate schedules built and evaluated: an MCTS scheduling method, preregistered as the main arm; a randomized-restart list scheduler with no search tree as the control; and a textbook-UCT MCTS arm as a registered follow-up on the same domain. The analysis plan was locked before data collection; an independent checker confirmed all 768 confirmation-stage schedules valid with correct finish times. At the main budget of 2048 evaluated schedules, MCTS is \textbf{2.88\% worse} than the restarts (95\% CI $[-3.62\%,-2.15\%]$; the 72 of 96 held-out instances with positive baseline-relative headroom). The deficit is not budget starvation: on the separate exploration pool restarts self-improve by 1.85\% where MCTS gains 0.24\%, and 700 of 720 of MCTS's first results are merely the warm start it was handed. The studied selection rule locks deterministically onto the first-expanded child, keeping its trials correlated while restarts draw gains from the tails; the causal link to makespan remains an inference. A structural map locates where the tree stops paying: restarts win on sparse irregular workflows, no difference is detectable on regular layered ones, and dense ones leave no room to improve. Swapping the rule for textbook UCT removes most of the deficit ($-0.26\%$, 95\% CI $[-0.71\%,+0.23\%]$) without reversing its sign. Results are confined to a controlled synthetic domain ($N\in\{8,16,32\}$ tasks, 2--8 processors, budgets up to 2048); we make no claim about larger budgets.

\end{abstract}

\section{Introduction}
\label{sec:intro}
A Monte Carlo tree search (MCTS) builds a search tree incrementally and uses statistics collected at each node to decide where to search next; it is a recurring proposal for scheduling directed acyclic graph (DAG) workflows. The appeal is clear: at each step the ready tasks and available processors form a sequential decision, and tree search is a natural tool for such decisions. The literature reports many positive results for MCTS schedulers and their variants \citep{kung2022,gnn_heatmap_mcts_2025}.

These evaluations, however, rarely include the one comparison that could refute the proposal: an equal-budget search that does not use a tree. Its absence leaves ``MCTS works'' ambiguous among three unrelated explanations---that the tree structure helps, that repeated randomized sampling helps, or that the warm-start heuristic, a good initial schedule handed to the search, seeds it. An equal-budget, tree-free control removes the second explanation from the list, which is the largest step available to us; it does not isolate the tree as a cause, because our MCTS arm additionally carries a warm start that the control lacks (Section~\ref{sec:method}). What we compare are therefore two complete procedures, and we ask a preregistered, falsifiable question:
\begin{quote}
\textbf{RQ1.} In a controlled synthetic domain, and under an identical search budget, does the MCTS scheduler outperform randomized-restart list scheduling?
\end{quote}
Our answer is negative. The reusable part is \emph{how} we answer it---the protocol.

The contributions split into what is learned and what is usable.
\begin{enumerate}
\item \textbf{A falsifiable protocol for budgeted search in scheduling.} We pair MCTS against a randomized-restart control that consumes an identical budget, one evaluated candidate schedule per restart, and fix the evaluation in advance: the analysis plan, covering instances, tests, and the reported effect, is locked before data collection; test instances are used once for confirmatory inference; the target quantity is declared beforehand as the mean improvement over instances with positive baseline-relative headroom ($H>\varepsilon$, defined as instances whose heuristic baseline our lower bound does not already certify as optimal); and an independent verifier checks the final schedule of every confirmation-stage run.
\item \textbf{The preregistered negative result.} At the primary budget of 2048 evaluated schedules, the main MCTS configuration is 2.88\% worse than the control (95\% confidence interval $[-3.62\%,-2.15\%]$, $n=72$). Across four structural strata with multiple-comparison correction, two are statistically significant, both favoring the control. A registered follow-up swaps that selection rule for textbook UCT in the same domain and removes most of this deficit---to $-0.26\%$, 95\% CI $[-0.71\%,+0.23\%]$---without reversing its sign (Appendix~\ref{app:standard-uct}): the magnitude belongs to the rule; the ordering does not, and the rule-robust comparison is itself a null, since the interval crosses zero and neither method is shown ahead.
\item \textbf{A reusable failure taxonomy.} Three patterns recur across an exploration stage and a held-out confirmation stage. \emph{Budget-monotone reversal:} MCTS leads at the smallest budget, by 1.14\% in exploration and 1.87\% on the confirmation domain, but falls behind once the budget reaches 8, respectively 4. \emph{Structural divergence:} across our four families, whether restarts win tracks edge density and irregularity---a descriptive pattern, not an isolated mechanism, since family identity is confounded with every structural attribute, namely density, irregularity and depth, at once. Restarts win on sparse, irregular DAGs; the two methods show no detectable difference on regular layered DAGs, an underpowered stratum; dense DAGs leave no room for either method to win; and exploration-stage chains point the same way, outside the confirmation domain. \emph{Saturation:} by the primary budget the median improvement per added budget has fallen to zero for both methods, so within the tested ladder the reversal is not explained by stopping early; budgets beyond 2048 are untested.
\end{enumerate}
We do not claim that MCTS cannot help scheduling, that this implementation stands for all MCTS schedulers, or that these synthetic results carry over to production workflows. What we show is one carefully controlled instance of the question, answered no: in the single setting where we made the comparison falsifiable, the MCTS procedure did not pay for its cost, and where it falls behind is identifiable. For practitioners whose workflows look like ours---sparse, irregular, small---the taxonomy gives a starting guess of when restarts suffice; it should be revalidated on real workloads before use. For researchers proposing MCTS schedulers, the protocol is the reusable part: it can make their next positive claim falsifiable.

\section{Related work}
\label{sec:related}
\textbf{List scheduling and classical baselines.} HEFT and CPOP \citep{topcuoglu2002} and PEFT/OCT \citep{arabnejad2013} define the list-scheduling baseline family: greedy processor assignment along a priority order. Our control arm is deliberately built from this family's components---minimum earliest-finish-time processor selection with deterministic tie-breaks---plus randomization, so the two arms differ in \emph{search strategy} rather than in the quality of the underlying placement heuristic; the warm-start asymmetry is stated in Section~\ref{sec:limitations} and is not controlled away. The wider DAG-scheduling literature spans list-scheduling heuristics and surveys \citep{kwok1999,sih1993,wu1990,kwokbench1999}, lookahead HEFT variants \citep{bittencourt2010}, optimal two-processor scheduling \citep{coffman1972}, and community workflow suites \citep{coleman2022}; scheduling hardness and worst-case bounds are classical \citep{graham1969,ullman1975}.

\textbf{Anytime and randomized-restart search.} Our control arm is grounded in the anytime and randomized-restart tradition: randomized restarts as a boosting strategy for combinatorial search \citep{gomes1998,gomes2000}, optimal restart schedules for Las Vegas algorithms \citep{luby1993}, anytime algorithms \citep{zilberstein1996}, iterated local search \citep{lourenco2003}, and greedy randomized adaptive search \citep{feo1995}. This is also where our mechanism lives: independent restarts profit from the tail of the schedule distribution, which connects to the max-$K$-armed / extreme bandit literature, where the objective is the best single draw rather than the mean \citep{cicirello2005,carpentier2014,achab2017,nishihara2016}. That framing explains why an i.i.d.\ restart sampler, which keeps drawing fresh extremes, can outperform a correlated, low-diversity tree search under an equal evaluation budget.

\textbf{MCTS for workflow scheduling.} Direct prior work includes Kung et al.~\citep{kung2022} and a GNN-heatmap-augmented MCTS \citep{gnn_heatmap_mcts_2025}; we cite these as the motivating family, and our study is an evaluation rather than a new method. A Chinese patent (CN109857532B) \citep{cn109857532b} combines critical-path (CPOP) information with MCTS DAG scheduling; our work is an evaluation study, not a method proposal.

\textbf{MCTS foundations and learned variants.} The studied selection rule should be read against canonical MCTS. UCT introduced UCB1-style tree selection \citep{kocsis2006}; Monte-Carlo evaluation with selectivity and backup is due to Coulom~\citep{coulom2006}; rapid action-value estimation (RAVE) \citep{gelly2011} and progressive strategies \citep{chaslot2008} are standard refinements; broad surveys are given by Browne et al.~\citep{browne2012} and Swiechowski et al.~\citep{swiechowski2023}. The exploration constant and randomized multiplier in Eq.~\eqref{eq:score} are far from these canonical forms, which is why we report it verbatim rather than as representative. Finite-time bandit analysis underlies the exploration term \citep{auer2002}, and MCTS with learned models produced landmark results in Go \citep{silver2016,silver2017}. A large modern line couples MCTS with learned value/policy models: the MuZero family and sampled variants for large action spaces \citep{schrittwieser2021}, Gumbel-based policy improvement \citep{danihelka2022}, multi-agent extensions \citep{hao2024}, and inference-time tree search for language models \citep{ugadiarov2026}. Our study targets a heuristic-seeded, model-free scheduler, so these are context rather than baselines; we make no claim about learned MCTS.

\textbf{Evaluation methodology.} Our protocol follows the operations-research tradition of designed computational experiments: reporting standards for heuristic comparison \citep{barr1995}, the call to test heuristics against meaningful baselines rather than convenient ones \citep{hooker1995}, and the critique of novelty claims built on uncontrolled comparisons \citep{sorensen2015}. On top of that tradition we adopt preregistration with hypothesis--confirmation separation \citep{nosek2018}. Our held-out set is used once for confirmatory inference; the append-only follow-ups in Appendices~\ref{app:standard-uct}--\ref{app:selection} reuse the same instances only as non-confirmatory evidence. We cite the reusable-holdout literature \citep{dwork2015} with an explicit downgrade, since our confirmatory holdout is one-shot rather than reusable. Broader concerns about reproducibility and evaluation rigor in machine learning motivate the preregistered, verifier-backed protocol we adopt \citep{henderson2018,gundersen2018}.

\textbf{Datasets and artifacts.} We use archived Pegasus workflow instances from WfCommons \citep{wfcommons_pegasus_instances} as background reference material (Zenodo DOI 10.5281/zenodo.7865734, other-open license); all quantitative results in this paper are on our synthetic domain, whose generator is deterministic and committed. The companion artifacts for this paper are archived with DOIs: the benchmark suite, covering instance generators, scheduler arms, independent verifier, and result tables, at \mbox{10.5281/zenodo.22929494}, and the evaluation audit protocol, covering the placebo standard and instrumentation contract, at \mbox{10.5281/zenodo.22930513}.

\section*{Notation}
\begin{center}
\footnotesize
\begin{tabular}{l@{\hspace{2em}}p{0.78\linewidth}}
\toprule
Symbol & Meaning \\
\midrule
$B$ & evaluated-search budget (candidate schedules built and evaluated) \\
$N$ & number of tasks in a DAG instance \\
$P$ & number of (equal-speed) processors, $P\in\{2,4,8\}$ \\
$\mathrm{CCR}$ & communication-to-computation ratio \\
$H$ & improvement headroom, $(M_0-L)/M_0$ (unrelated to H-arm) \\
$M_0$ & per-instance best of a four-heuristic portfolio (HEFT, CPOP, PEFT, OCT) \\
$L$ & lower bound, $\max(L_{\mathrm{work}},L_{\mathrm{cp}})$ \\
$\varepsilon$ & headroom threshold ($0$) \\
$\varepsilon_{\mathrm{inc}}$ & incumbent-improvement tolerance ($10^{-9}$) \\
$\Delta$ & per-instance relative gap, $(\mathrm{control}-\mathrm{mcts})/\mathrm{control}$; $\Delta>0$ favors MCTS \\
$F$ & finish time ($F_{\mathrm{dispatch}}\ge F_{\mathrm{precedence}}$) \\
$n$ & conditional-domain sample size (72 of 96) \\
$N_v,N_{v_i}$ & visit counts of node $v$ and child $v_i$ \\
$Q_{v_i}$ & makespan-derived value backed up at child $v_i$ \\
$\text{default\_Q}$ & per-instance constant (total work plus outgoing communication) \\
$\text{rng}$ & fresh 32-bit pseudorandom draw, one per scoring \\
$c$ & UCB1 exploration constant ($c=0.7$ in Appendix~\ref{app:standard-uct}) \\
W/T/L & per-instance win/tie/loss (from the MCTS side) \\
HL & Hodges--Lehmann location estimate \\
MDE, $d_z$ & minimum detectable effect; paired effect size \\
H/R/T & original implementation; reimplementation; topology-aware main arm \\
\bottomrule
\end{tabular}
\end{center}

\section{Method}
\label{sec:method}
\subsection{The scheduler under study and two communication models}
We study three arms (arm: algorithm configuration) of one scheduler: our own earlier implementation, which predates this study and is kept unmodified here (the H-arm), a deterministic reimplementation proven equivalent to it within bounds (the R-arm), and a topology-aware rewrite that forms the main arm (the T-arm); these arm names carry no relation to the headroom $H$ of Eq.~\eqref{eq:headroom}. At each node the scheduler scores every child action and selects the argmax of
\begin{equation}
\mathrm{score}(v_i)=\frac{\text{default\_Q}}{Q_{v_i}}\cdot(\text{rng}\bmod 5)+10^{5}\cdot\sqrt{\frac{\ln N_v}{N_{v_i}}},
\label{eq:score}
\end{equation}
where $N_v$ and $N_{v_i}$ are the visit counts of a node and a child, and $Q_{v_i}$ is the makespan-derived value backed up at child $v_i$, with makespan denoting the job's finish time. The constant $\text{default\_Q}$ is per-instance: the total task execution time plus the total outgoing communication time. It also initializes a freshly expanded node, and is therefore identical across siblings. The term $\text{rng}$ is a fresh 32-bit draw per scoring, and $10^{5}$ is the exploration constant as given in that implementation. An unexpanded action is scored \texttt{INT\_MAX} so it is expanded first. We report Eq.~\eqref{eq:score} verbatim as the original, nonstandard rule, and we neither ``fix'' nor tune it. We have since measured the selection distribution directly (Appendix~\ref{app:selection}): in the configuration we report, the value channel is inert---a zero-initialised minimum makespan makes every child's value term diverge---so under strict-greater comparison the search locks deterministically onto the first-expanded child rather than spreading near-uniformly; the large constant and the $(\text{rng}\bmod 5)$ multiplier dominate the score only once the value channel is repaired.

\paragraph{The main arm as an anytime search.} A search state $s$ carries the scheduled set, each task's processor and finish time, per-processor availability, the ready set $\mathrm{Ready}(s)=\{i:\text{all predecessors of }i\text{ are scheduled}\}$, and the action prefix; the legal actions are $A(s)=\mathrm{Ready}(s)\times\text{processors}$. Taking action $(i,p)$ sets $\mathrm{start}(i,p)=\max(\mathrm{avail}(p),\ \max_{j\in\mathrm{pred}(i)}(\mathrm{finish}(j)+\mathrm{comm}(j,i)))$ and $\mathrm{finish}(i,p)=\mathrm{start}(i,p)+w(i,p)$. One search descends from the root by Eq.~\eqref{eq:score}, expands one node, completes the schedule with an earliest-finish-time rollout, and backs up visit counts and makespan statistics along the path; the incumbent is the best complete schedule seen so far. The budget $B$ counts \emph{evaluated} searches, and a warm-start replay that re-runs a full rollout consumes one of them: it is counted against $B$ but flagged as a duplicate rather than credited as new search---which is why the arm's $B=1$ result is usually exactly its warm start. The main arm is seeded with a CPOP warm start, whose asymmetry we return to in Section~\ref{sec:limitations}.

\paragraph{Scope of the H/R/T correspondence.} R is a from-scratch reimplementation of the historical scheduler H; the two are shown to agree in a \emph{bounded, correctness sense} on a restricted representable slice, namely zero-communication edges, homogeneous unit speed, non-negative integer costs with cumulative $\text{default\_Q}\le\texttt{INT32\_MAX}$, and bidirectionally consistent DAGs, with an archived per-cell parity receipt in the artifact bundle. This is behavioral parity, not performance parity: we make no runtime-equivalence claim, and the historical performance question is out of scope here. T is the topology-aware rewrite that carries the precedence-aware communication model and forms the main arm.

Two communication models appear in the literature and in the code. The \textbf{precedence-aware model} sets each task's start to the latest parent-arrival time, so communication respects precedence; the \textbf{dispatch-delay model} instead charges an edge delay at dispatch. Writing $F$ for the resulting finish times, $F_{\mathrm{dispatch}}\ge F_{\mathrm{precedence}}$ holds as a precondition, because the dispatch-delay model charges communication no earlier than the precedence-aware parent-arrival model does. The main table compares arms \emph{within the precedence-aware model}, so the MCTS and control arms are scored under identical semantics and pairing is valid; data from the original selection formula are reported separately (Appendix~\ref{app:dyadic-formula}) and never shown as the main arm.

\subsection{The control arm}
The control arm is a randomized-restart list scheduler. Each restart starts from an empty schedule and, at every step, draws one task uniformly at random from the \emph{current} ready set, implemented as a Fisher--Yates shuffle of that set with the front taken, and places it on the processor of minimum earliest-finish time with deterministic tie-breaks---matching the processor rule the MCTS arm uses. It is thus a random list schedule, not one up-front permutation of all tasks. The incumbent is replaced only on strict improvement, with $\varepsilon_{\mathrm{inc}}=10^{-9}$, and \textbf{each restart consumes exactly one evaluated search}, so a budget of $B$ means $B$ evaluated schedules for either arm. The control arm entered the preregistration only after passing four correctness gates: independent-verifier PASS, budget conservation, incumbent monotonicity, and same-seed replay determinism.

\subsection{Instances and improvement headroom}
The generator spans six structural families, chain, fork-join, layered, critical-path-heavy, dense, and sparse-wide, with $N\in\{8,\dots,128\}$, three processor counts $P=2,4,8$ at equal speed, and communication-to-computation ratio $\mathrm{CCR}\in\{0,0.5,2,8\}$; it replays byte-identically under a fixed seed and covers two cost domains in the exploration stage: dyadic, meaning costs drawn from powers of two, and integer.

The preregistered held-out, confirmation domain narrows this grid. It is the four-family subset, dropping chain and critical-path-heavy, with $N\in\{8,16,32\}$, $P\in\{2,4,8\}$ and $\mathrm{CCR}\in\{0.5,2\}$: $96=4\ \text{families}\times2\ \text{CCR levels}\times12\ \text{instance seeds}$, with the 12 seeds in each family--CCR block cycling through the nine $(N,P)$ combinations, so that grid is covered but not balanced within a block. Improvement headroom is
\begin{equation}
H=\frac{M_0-L}{M_0},\qquad L=\max(L_{\mathrm{work}},L_{\mathrm{cp}}),
\label{eq:headroom}
\end{equation}
where $M_0$ is the per-instance best of the four-heuristic portfolio (HEFT, CPOP, PEFT, OCT) and $L$ is a lower bound taken as the larger of a work bound and a critical-path bound, not their sum; the critical-path bound ignores communication. The threshold $\varepsilon=0$ was fixed before execution. Filtering on $H$ conditions on baseline-relative headroom---a preregistered selection-bias disclosure---so we also report the unfiltered distribution in Appendix~\ref{app:unfiltered}.

\subsection{Preregistration and statistics}
The confirmation-stage manifest was frozen before data collection: 96 held-out instances $\times$ 8 arm--seed configurations, comprising two registered MCTS variants $\times$ two search seeds plus four control seeds, $\times$ 12 budgets $=9{,}216$ rows; an anonymized registration record is available to reviewers. The statistical unit is the DAG instance: repeated seeds are an algorithm configuration rather than independent replicates, so each arm's per-instance value is its mean over the registered seeds, and the sample size $n=96$ was fixed in advance. The registered estimand is the conditional mean of Eq.~\eqref{eq:estimand}, reported with a bootstrap percentile confidence interval from $10^4$ instance-level resamples; the registered confirmatory test is a two-sided paired sign-flip permutation test within each family stratum, with Wilcoxon as reference, Holm-corrected across the four strata, with Hodges--Lehmann (HL) location estimates. These are two distinct objects rather than one: the pooled $-2.88\%$ is the descriptive main estimand, and its uncertainty is expressed by the bootstrap interval, whereas the confirmatory decision is made per family stratum by the Holm-corrected permutation test rather than on the pooled number itself; we report both and do not read the pooled interval as a significance test. 

The Holm family is the four preregistered family strata as declared, which matters once the headroom filter empties one of them: the dense stratum contributes no test and enters the correction with $p=1$, so the correction divides over $m=4$ rather than over the three estimable strata---a conservative choice fixed by the registered analysis code, not chosen after the fact. The independent verifier is a separate implementation that consumes only an instance definition and an emitted schedule; it shares no scheduling code with either arm and takes no part in generating candidates. A void-run protocol separates harness crashes from scientific outcomes; there were zero void runs at the confirmation stage. Upon acceptance we will release the registration record, the instance generator, the verifier, and the per-cell parity receipts. The target quantity is the conditional mean
\begin{equation}
\mathbb{E}\!\left[\,\Delta\mid H>\varepsilon,\ \text{domain}\,\right],\qquad
\Delta=\frac{\mathrm{control}-\mathrm{mcts}}{\mathrm{control}}\ \text{per instance},
\label{eq:estimand}
\end{equation}
with $\Delta>0$ favoring MCTS.
\textbf{Preregistered arms.} The frozen manifest declared two MCTS arms---the main arm, using the precedence-aware model, and an auxiliary arm---and asserted that both ran under the control arm's communication semantics. That assertion was wrong for the auxiliary arm, which is a dispatch-delay arm, so a paired $\Delta$ against the precedence-aware control would cross communication models and would measure a change of units rather than a difference in scheduling quality. The mismatch was identified while the confirmation-stage executor was being written---its arm-aware verifier already applies dispatch-delay semantics to that arm, and it was committed before the run produced any outcome---so the exclusion criterion is mechanical and was fixed before any result existed. The auxiliary arm was nonetheless executed exactly as registered: its rows are part of the 9{,}216 and its within-model statistics are in the released artifact. The main table reports the main arm only, and we log this as a registered erratum.

\section{Results}
\label{sec:results}
\textbf{Data.} Stage 1, the exploration stage, is research-only and excluded from the main table; it comprises 77{,}760 rows across nine arms: seven MCTS variants plus two controls, one on the integer-cost cells of the exploration matrix and one on its dyadic-cost cells, the latter being the \emph{dyadic control} referred to below. Under the registered no-deletion rule, 3{,}351 adverse rows---outside the target domain or unfavourable to our own hypothesis---were retained rather than dropped. Stage 2, the main table, comprises 9{,}216 arm rows with zero void runs (a void run is a harness failure rather than a scientific outcome); the 96 recorded runs beyond the arm rows are the one-per-instance Portfolio-4 baseline and headroom computation. All 768 of 768 final schedules at the primary budget pass the independent verifier. Baselines are the per-instance Portfolio-4 best (HEFT/CPOP/PEFT/OCT).

\begin{figure}[t]
\centering
\includegraphics[width=0.66\linewidth]{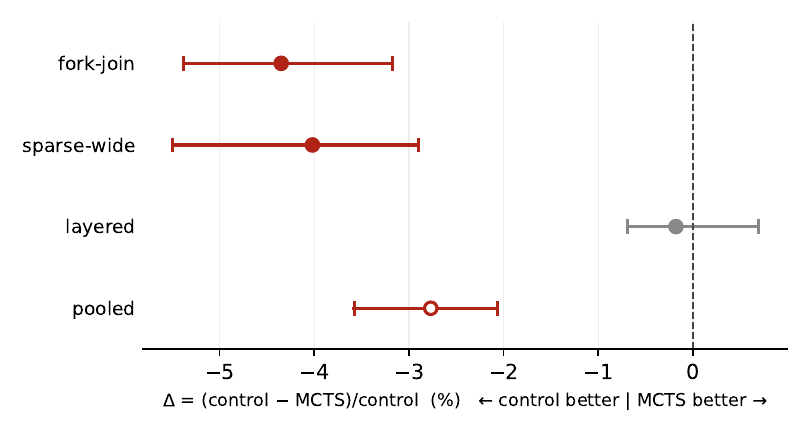}
\caption{Per-stratum effect, shown as the Hodges--Lehmann estimate with 95\% CI, for the main comparison. Fork-join and sparse-wide exclude zero on the control's side; layered spans zero; the pooled estimate is $-2.88\%$. Negative favors the control.}
\label{fig:forest}
\end{figure}

\textbf{Main result.} On the conditional domain, $\mathbb{E}[\Delta\mid H>\varepsilon]=-0.0288$ with 95\% CI $[-0.0362,-0.0215]$ and $n=72$; win/tie/loss $=7/7/58$. Of the 96 held-out instances, 72 pass the preregistered $H>\varepsilon$ filter; the 24 that do not are all dense-family cells with $H=0$, where the portfolio best already meets the lower bound, so the conditional domain excludes the dense family by construction. After Holm correction, two strata are significant, both favoring the control: fork-join, with $p\approx1.2\times10^{-7}$, and sparse-wide, with $p\approx4.8\times10^{-7}$; the layered stratum is not significant, with Holm $p=1.0$ and raw permutation $p=0.9613$. Table~\ref{tab:strata} gives the per-stratum statistics, and Figure~\ref{fig:forest} shows the same estimates with their confidence intervals.

\begin{table}[t]
\centering
\caption{Per-stratum statistics for the main comparison (main MCTS arm vs.\ control, precedence-aware model, $B=2048$). W/T/L $=$ MCTS win/tie/loss per instance; HL $=$ Hodges--Lehmann estimate.}
\label{tab:strata}
\begin{tabular}{lrrrrr}
\toprule
Stratum & $n$ & W/T/L & mean $\Delta$ & HL & Holm sig. \\
\midrule
fork-join   & 24 & 0/0/24 & $-0.0436$ & $-0.0435$ & yes ($p\approx1.2\times10^{-7}$) \\
sparse-wide & 24 & 0/2/22 & $-0.0426$ & $-0.0402$ & yes ($p\approx4.8\times10^{-7}$) \\
layered     & 24 & 7/5/12 & $-0.0002$ & $-0.0018$ & no ($p=1.0$) \\
dense       & 0  & ---    & ---       & ---       & --- ($n=0$) \\
pooled      & 72 & 7/7/58 & $-0.0288$ & $-0.0277$ & --- \\
\bottomrule
\end{tabular}
\end{table}

\begin{figure}[t]
\centering
\includegraphics[width=0.72\linewidth]{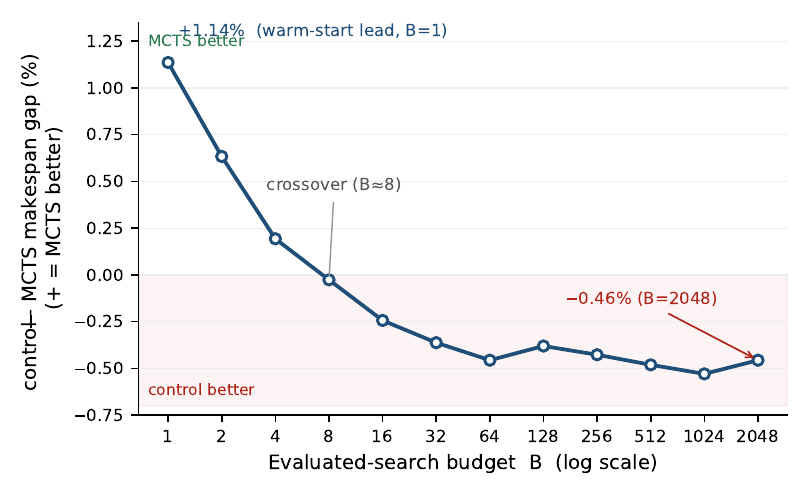}
\caption{Budget-monotone reversal (exploration stage, 720 cell--seed pairs pooled). The paired gap between the MCTS arm and the control, $\Delta=(\text{control}-\text{MCTS})/\text{control}$ in the sign convention of Eq.~\eqref{eq:estimand}, averaged per budget, starts at $+1.14\%$, with MCTS ahead almost entirely as a warm-start effect, at $B=1$, crosses zero near $B=8$, and settles at $-0.46\%$ by $B=2048$. Positive values favor MCTS. Budgets beyond 2048 are not tested.}
\label{fig:reversal}
\end{figure}

\begin{figure}[t]
\centering
\includegraphics[width=0.72\linewidth]{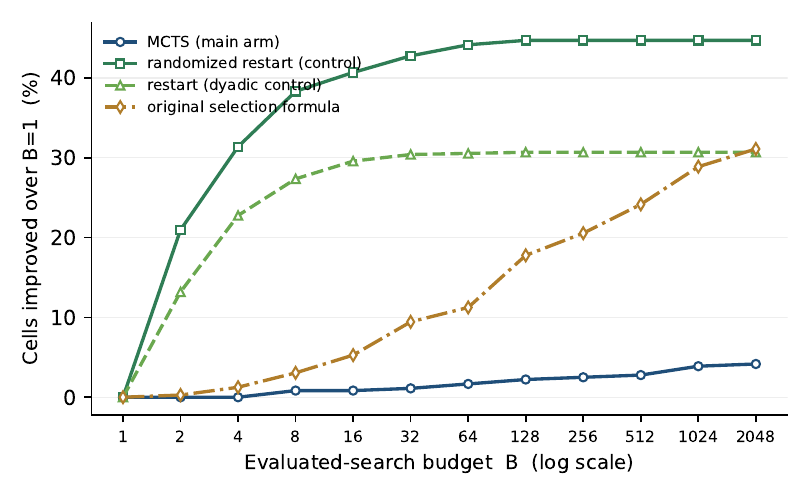}
\caption{Fraction of cells that improve over their $B=1$ result, versus budget (exploration stage, per arm). The MCTS arm barely rises, with about 4\% of cells improving even at $B=2048$, whereas the randomized-restart control reaches roughly 45\%; the dyadic control, which is the same control arm on the exploration matrix's dyadic-cost cells, and the original selection formula of Eq.~\eqref{eq:score} fall in between. More budget buys the tree almost nothing.}
\label{fig:anytime}
\end{figure}

\textbf{Is the deficit a budget artifact?} Three exploration-stage checks say no, within the tested ladder; complete exclusion at larger budgets remains open work. First, the MCTS anytime curve is \emph{flatter} than the control's: self-improvement from $B=1$ to $B=2048$ is 0.24\% for MCTS versus 1.85\% for the control, and the paired gap itself reverses from $+1.14\%$ to $-0.46\%$ over the same range (Figure~\ref{fig:reversal}). The confirmation domain shows the same reversal, earlier and larger: the paired gap there is $+1.87\%$ at $B=1$ and has already turned negative by $B=4$, reaching $-2.88\%$ at the primary budget. The endpoint magnitudes differ between the two stages because the pools differ---the exploration pool mixes in families that tie exactly and the dyadic cost domain, whereas the confirmation domain is restricted to four families of which two are headroom-rich---so we treat only the confirmation figure as the result. Second, MCTS has plateaued by $B=2048$: for the MCTS arm the median per-cell gain over the final budget doubling, from 1024 to 2048, is zero, and all but two cells gain less than 0.06\%; one of the two exceptions is a large layered gain. The control has not fully plateaued---a minority of its cells still gain under a percent, which only widens its lead---so within the tested ladder the reversal is not explained by stopping too early (Figure~\ref{fig:anytime}). Third, we found no sub-domain in which more budget closes the gap; this is a search that failed to find one, not a proof that none exists beyond $B=2048$. At $n=72$ the minimum detectable effect, with paired SD 3.23\%, $\alpha=0.05$, and power 0.8, is 1.07\%, well below the observed 2.88\%; the paired effect size is $d_z\approx-0.9$, with the bootstrap CI excluding zero.

\textbf{Mechanism, with selection measured and makespan attribution still inferred.} The paired deficit decomposes into a start-point difference and a learning-rate difference. At $B=1$, the MCTS arm's $+1.14\%$ lead is almost entirely its CPOP warm start: on 720 exploration instances, 700 of the $B=1$ incumbents \emph{are} the warm start---the tree has not searched yet. By $B=2048$ the control overtakes with a 7.8$\times$ larger endpoint self-improvement, 1.85\% versus 0.24\%, from a worse start; the two percentages are measured from different starting points and are not commensurable learning rates, which is why the quantity we report as the result is the endpoint gap rather than their ratio.

We now measure the selection distribution directly (Appendix~\ref{app:selection}); its causal role in the makespan deficit remains an inference rather than a measured quantity. In the configuration we report the value channel is inert---a zero-initialised minimum makespan makes every child's value term diverge---so under strict-greater comparison the search locks deterministically onto the first-expanded child, with mean top-1 visit share $0.86$ rising toward $1.0$ as the root fills, and normalized selection entropy $0.29$ against $1.00$ for a value-repaired canary. Successive tree samples stay correlated and carry little new information not because the tree explores near-uniformly but because it barely explores at all, whereas independent restarts keep drawing i.i.d.\ samples whose extremes pay off. This value-blind, first-child lock-in is the same degeneracy family as the rank-$0$ selection signature of the original implementation, whose value is $0.82$; throughout, ``original implementation'' denotes the earlier MCTS scheduler we wrote before this study and reproduce here verbatim. The degeneracy is a tree with depth but no breadth, the opposite of one that ``concentrates too much'' by over-exploiting a well-estimated value.

The original selection formula of Eq.~\eqref{eq:score} was additionally run in the exploration stage's dyadic-cost domain; those raw counts are reported separately in Appendix~\ref{app:dyadic-formula} and are never shown as the main arm.

\section{Failure taxonomy and selector}
\label{sec:taxonomy}
\begin{figure}[t]
\centering
\includegraphics[width=0.66\linewidth]{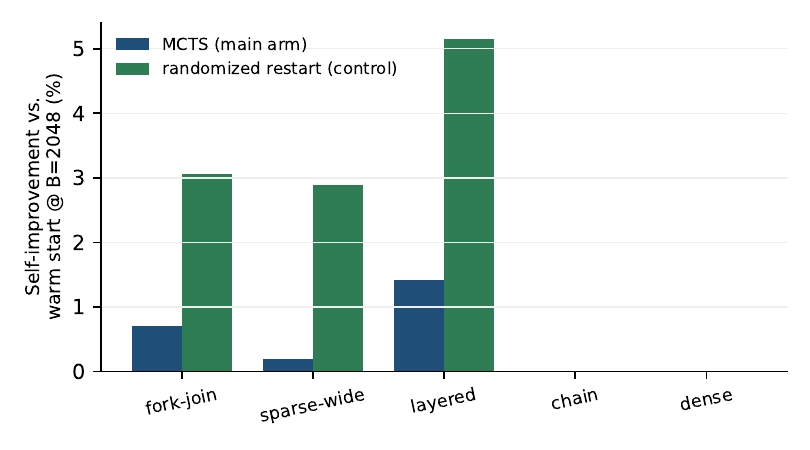}
\caption{Self-improvement over the warm start at $B=2048$, by family. The control extracts $3$--$4\times$ more than MCTS on every family that has any headroom, namely fork-join, sparse-wide, and layered; chains and dense DAGs have none. The $3$--$4\times$ figure is per family; pooled across all families the ratio is the larger $7.8\times$ reported in the main text, because pooling aggregates the per-family gains.}
\label{fig:family}
\end{figure}

Three reusable patterns locate where the tree stops paying; this is contribution (iii). \emph{(a) Budget-monotone reversal:} the MCTS advantage decays with budget and reverses by $B\ge8$ in the exploration scan, and by $B\ge4$ on the confirmation domain. Where the crossing falls is a property of the seeding protocol, since only the MCTS arm is warm-started; the budget-robust content of this pattern is the difference in slope, not the crossing budget. \emph{(b) Structural divergence:} across our four families, whether restarts win tracks edge density and irregularity---the losing families, sparse-wide and fork-join, are low-density and irregular, while the layered family, which consists of three complete bipartite layers at depth 3 with $E/N\approx3.4$, shows no detectable difference. Family identity is confounded with density, irregularity and depth at once, so this describes which cells lose rather than isolating a causal effect of density. Dense DAGs leave no headroom to win, and chains behave alike in exploration-stage data (Figure~\ref{fig:family}). \emph{(c) Saturation:} the median per-cell slope over the final budget doubling is zero on both arms, so within the tested ladder the reversal is not explained by stopping early.

Mapped to the preregistered strata with evidence levels: sparse-wide and fork-join are the \textbf{restart-favoring} region, confirmed at the confirmation stage and significant after Holm; layered is the \textbf{no-detectable-difference} region, established at the confirmation stage with HL CI $[-0.7\%,+0.7\%]$ and stratum MDE $\approx1.2\%$---that is, underpowered rather than proven equal; dense DAGs are a \textbf{ceiling} region rather than a search outcome, since all 24 held-out dense instances have $H=0$, so our lower bound already certifies the Portfolio-4 baseline as optimal there, and separately the two arms are observed to tie on these cells (Appendix~\ref{app:unfiltered}); and chains show no difference in exploration-stage data only---the chain family lies outside the confirmation domain, so this cell of the taxonomy is unconfirmed and untested. \emph{Selector.} For practitioners under equal budget: use randomized restarts outright on low-density irregular structures; expect no detectable difference on regular bipartite structures; and where the heuristic baseline already meets the lower bound, as on our dense DAGs, there is no headroom for either method to exploit and on our cells the two arms tie. For researchers, the transferable part is the protocol---an equal-budget control, an independent verifier, and a conditional estimand fixed in advance.

\section{Limitations}
\label{sec:limitations}
\textbf{Domain boundary, primary.} All conclusions are confined to the controlled synthetic domain, with integer costs and a unit-speed executor contract. We do not extrapolate to production workflows; heterogeneous-placement criteria are structurally unsatisfiable in this domain, and migrating there is registered as open work.

\textbf{Asymmetric warm start.} The main-table MCTS arm carries a CPOP warm start that the control arm lacks. The $B=1$ lead is almost entirely that warm start; the $B=2048$ deficit is the control achieving a 7.8$\times$ larger endpoint self-improvement from a worse start. We state the two effects separately wherever both appear. The asymmetry is conservative for the conclusion we draw, which we verify directly rather than assume. Under a preregistered symmetric protocol that hands the control the same CPOP warm start, the deficit persists and, if anything, widens, from $-2.88\%$ to $-2.93\%$, while the crossing budget of Figure~\ref{fig:reversal} moves; at the primary budget the warm start is worth a mean $+0.05\%$ to the control (Appendix~\ref{app:symmetric-warmstart}).

We therefore do not rest this on an a-priori argument that symmetrizing could only widen the gap. Warm start is also not the arms' only difference: tree structure, and the tree's freedom over processor placement, remain uncon\-trolled. This experiment addresses the warm-start asymmetry specifically rather than isolating the tree. It is reported as append-only evidence and modifies no frozen confirmation number.

\textbf{Layered-family density cap.} The layered family is a three-layer complete bipartite graph with $E=O(N^2)$ growth; the main-table instances, with $N\in\{8,16,32\}$, have $E/N\in[1.0,6.2]$, and the $N=32$ value of 6.2 slightly exceeds the real-workflow band, which has $E/N\in[2,5]$; this is a genuine consequence of the family's construction, not a generator artifact. The true artifact, $E/N\approx4444$, appears only in the diagnostic ladder at $N\ge10^3$ and is outside the main table.

\textbf{Scope of comparison.} Our claim is a preregistered equal-budget comparison against the list-scheduling family, HEFT, CPOP, PEFT, and OCT, not a comparison against state-of-the-art learning-guided schedulers; extending the protocol to learning-guided baselines is future work.

\textbf{Is this just a broken MCTS?} Two results answer this question before it is asked---this is the question the registered follow-up arm of the abstract was designed for. Replacing the studied rule with canonical UCT in the same domain removes most of the deficit---from $-2.88\%$ to $-0.26\%$ (95\% CI $[-0.71\%,+0.23\%]$; Appendix~\ref{app:standard-uct})---so the \emph{magnitude} is rule-specific; yet the direction survives, since under an equal budget canonical UCT still does not overtake independent restarts. With that shield stated, the provenance question remains fair: the rule of Eq.~\eqref{eq:score} is nonstandard---a $10^{5}$ exploration constant and a randomized exploitation multiplier---and one might ask whether we have merely tested a broken implementation. We wrote that rule ourselves, for earlier scheduling work, and we report it here verbatim rather than quietly repairing it: it predates this study, it was not constructed for this comparison, and neither arm was tuned after the protocol was registered. Our own authorship of the rule is precisely why provenance cannot settle the question, so we answer it with two experiments. In separate clean-room work, documented in the released bundle, a textbook UCT written from the standard specification without consulting the original implementation, using a standard exploration constant and an earliest-finish-time warm start, did \emph{not} reproduce the selection degeneration, and produced genuine search improvement rather than restart-style gains. A preregistered, same-domain standard-UCT arm has since been run (Appendix~\ref{app:standard-uct}): the pooled 95\% CI $[-0.71\%,+0.23\%]$ places any UCT advantage at no more than $+0.23\%$ while leaving open a disadvantage as well. The honest reading of our headline number is therefore that it belongs to the original rule, while the rule-robust statement is the weaker one---no selection rule we tested beats equal-budget restarts in this domain. We do not claim that standard UCT is significantly worse than the control.

Three further registered limitations are stated here. Performance-budget gates---wall-clock and memory ceilings per arm---were registered but not executed, so our equal-budget claim covers evaluated schedules only. An earlier real-workflow signal from a Python-era prototype is deprecated rather than refuted: it is excluded from every number in this paper, and we make no claim either way about it. Finally, several exploration-stage sub-pools are small enough that we describe them qualitatively instead of testing them. A fourth registered item, the selection-artifact experiment, has since been executed (Appendix~\ref{app:selection}): it measures the selection distribution directly---the causal step from that distribution to the makespan deficit remains an inference---and, honoring the preregistered honest-reporting clause, reverses its predicted direction, since the reported rule concentrates its trials on the first-expanded child rather than spreading them near-uniformly.

\section{Conclusion}
\label{sec:conclusion}
We asked a simple question: given the same search effort, does an MCTS scheduler beat a method with no tree at all? On our controlled synthetic domain the answer is no. At the preregistered budget the MCTS arm finished 2.88\% behind a randomized-restart list scheduler, and its only early advantage was a warm start it was handed---not anything the tree found. Two selection rules under one locked protocol give the same answer: the production rule loses outright, and textbook UCT, run as a registered follow-up on the same domain, closes most of the gap but does not reverse it.

The finding is deliberately narrow: one domain, one implementation, budgets up to 2048. What travels is the method---an equal-budget control with no tree, an analysis plan locked before we saw the data, an independent checker on every final schedule, and a target measure fixed in advance. Together they turn ``does MCTS help here?'' from an untestable intuition into a question that can be answered, and here answered no.

For practitioners under a fixed search budget, the taxonomy says where a handful of random restarts is the better buy and where the choice is immaterial; on our domain it contains no region in which the tree pays for itself. For researchers, the next MCTS-scheduling proposal can borrow the protocol and state a claim that could, in principle, be proven wrong---which is the first thing a useful claim should be able to do.

\appendix
\section{Unfiltered distribution (preregistration obligation)}
\label{app:unfiltered}
Per the frozen estimand clause, we report the unfiltered distribution alongside the conditional one, on all 96 held-out cells, at $B=2048$ with arm means over seeds and $\Delta=(\mathrm{control}-\mathrm{MCTS})/\mathrm{control}$; see Table~\ref{tab:unfiltered}.

\begin{table}[!ht]
\centering
\caption{Unfiltered distribution on the full 96 held-out cells.}
\label{tab:unfiltered}
\begin{tabular}{lrrr}
\toprule
Family & $n$ & mean $\Delta$ & range \\
\midrule
fork-join   & 24 & $-4.36\%$ & $[-9.21\%,-0.52\%]$ \\
sparse-wide & 24 & $-4.26\%$ & $[-11.30\%,0.00\%]$ \\
layered     & 24 & $-0.02\%$ & $[-4.81\%,+6.81\%]$ \\
dense       & 24 & $0.00\%$  & $[0.00\%,0.00\%]$ \\
\textbf{all (unfiltered)} & \textbf{96} & \textbf{$-2.16\%$} & --- \\
\bottomrule
\end{tabular}
\end{table}

The unfiltered pooled mean, $-2.16\%$, is closer to zero than the conditional one, $-2.88\%$, because the 24 dense cells all have $H=0$, where the portfolio best equals the lower bound, and contribute exact ties. Filtering on $H$ conditions on baseline-relative headroom: the headroom distribution differs by CCR, with median $H=0.126$ at low CCR versus $0.390$ at high CCR and ranges $[0,0.42]$ and $[0,0.57]$, and high-CCR instances---where scheduling has the most room---are also where the control's mean advantage is largest in our cells. That last point is a descriptive reading of the per-cell table, not a CCR-stratified test, which we do not run.

\section{Original selection formula in the dyadic domain (exploration-stage raw counts)}
\label{app:dyadic-formula}
The rule of Eq.~\eqref{eq:score} was run in the exploration stage's dyadic-cost domain and reported separately, never as the main arm. It improves on its own start on 224 of 720 instances, versus 221 of 720 for the dyadic control---essentially level. These are raw counts that we do not test, and we read the three-instance difference as no advantage in either direction. Both counts are on the dyadic-cost exploration pool and are not directly comparable to the integer-cost improvement fraction of the main arm in Figure~\ref{fig:anytime}, which is a different cost domain and instance pool.

\section{Same-domain standard-UCT arm (registered follow-up, completed)}
\label{app:standard-uct}
Section~\ref{sec:limitations} registered an in-domain standard-UCT arm as follow-up work, to turn the ``broken MCTS?'' rebuttal from clean-room, indirect, into same-domain, direct, evidence. That arm has since been run under a preregistered protocol. We report it here as \emph{append-only} evidence: it introduces no change to any frozen confirmation-stage number, criterion, or conclusion in the main text.

\textbf{Protocol.} We replace the original selection rule of Eq.~\eqref{eq:score} with a textbook UCT/UCB1 selection, with exploration constant $c=0.7$, and hold everything else in the main arm fixed: the CPOP warm start, the earliest-finish-time rollout, the precedence-aware communication model, the single-unvisited-action-per-evaluated-search expansion, and the evaluated-search budget ladder. Rewards are normalized before they enter UCB1: a rollout's reward is $\min(1,\ \text{warm-start makespan}/\text{rollout makespan})\in(0,1]$ and the exploitation term is the mean of those rewards, so $c=0.7$ is applied on the $[0,1]$ scale for which it is the textbook default rather than to a raw makespan. Because the ratio is clipped at $1$, every rollout at least as good as the warm start receives the maximal reward $1$: the exploitation signal saturates above the warm-start makespan, so a schedule that beats the warm start earns no higher reward than one that merely ties it. This arm is therefore textbook UCT \emph{selection} with this specified clipped reward, not a claim that a single reward shape or exploration constant represents all of standard UCT; we report it as one concrete, canonical selection rule rather than as a universal repair. The arm is scored on the same held-out instances and by the same independent verifier as the main comparison. The formal run comprises $6{,}912$ rows, that is $96$ instances $\times$ 12 budgets $\times$ two UCT search seeds plus four control seeds, with zero verifier failures; the UCT side therefore carries two seeds against the control's four, making its per-instance mean the noisier of the two and widening the interval below. Because this arm reuses the same 96 held-out instances, it is a second use of a one-shot holdout: we report it as append-only evidence, its Holm correction is internal to this appendix, and we do not treat its $p$-values as confirmatory.

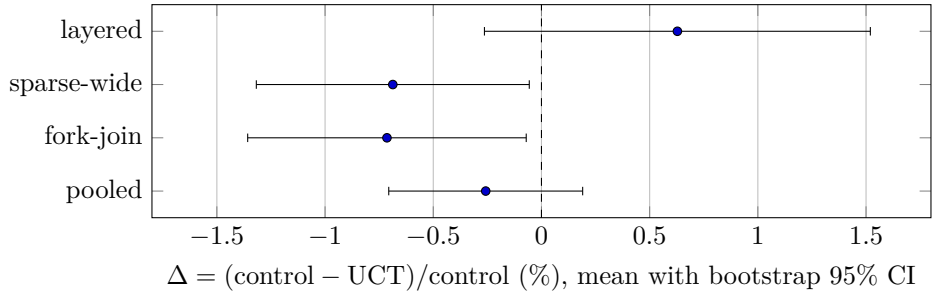
\begin{figure}[t]
\centering
\begin{tikzpicture}
\begin{axis}[
  width=0.72\linewidth, height=4.4cm,
  xlabel={$\Delta=(\text{control}-\text{UCT})/\text{control}$ (\%), mean with bootstrap 95\% CI},
  xmin=-1.8, xmax=1.8,
  ytick={1,2,3,4}, yticklabels={pooled,fork-join,sparse-wide,layered},
  ymin=0.5, ymax=4.5,
  xmajorgrids, tick align=inside, enlarge y limits=false,
]
\draw[dashed] (axis cs:0,0.5) -- (axis cs:0,4.5);
\addplot+[only marks, mark=*, black, mark size=1.6pt,
  error bars/.cd, x dir=both, x explicit]
 coordinates {
  (-0.2576,1) +- (0.4481,0.4868)
  (-0.7140,2) +- (0.6436,0.6323)
  (-0.6870,3) +- (0.6307,0.7065)
  (0.6282,4)  +- (0.8917,0.9753)
 };
\end{axis}
\end{tikzpicture}
\caption{Same-domain standard-UCT arm: per-stratum mean $\Delta$ with bootstrap 95\% CI, and the pooled estimate ($B=2048$, $H>\varepsilon$). Negative favors the control. The fork-join interval lies entirely below zero but does not survive Holm correction, with $p=0.174$; the sparse-wide, layered and pooled intervals cross zero; no stratum passes correction. The bootstrap intervals shown are not Holm-corrected whereas the $p$-values are, so an interval that excludes zero does not by itself imply significance after correction.}
\label{fig:uct-forest}
\end{figure}

\begin{figure}[t]
\centering
\begin{tikzpicture}
\begin{axis}[
  width=0.74\linewidth, height=5cm,
  xlabel={evaluated-search budget}, ylabel={mean makespan (lower is better)},
  xtick={0,1,2,3,4,5,6,7,8,9,10,11},
  xticklabels={1,2,4,8,16,32,64,128,256,512,1024,2048},
  x tick label style={font=\scriptsize,rotate=45,anchor=east},
  ymin=319.5, ymax=328, ymajorgrids,
  legend style={at={(0.98,0.98)},anchor=north east,font=\small},
  legend cell align=left,
]
\addplot[color=red!70!black, mark=*, mark size=1pt] coordinates {
 (0,324.2188)(1,324.2188)(2,324.0312)(3,323.8021)(4,323.2292)(5,323.0625)(6,322.7812)(7,322.0521)(8,321.4792)(9,321.2396)(10,321.0625)(11,320.7708)};
\addlegendentry{standard UCT}
\addplot[color=blue!60!black, mark=square*, mark size=1pt] coordinates {
 (0,327.4089)(1,325.5312)(2,324.0755)(3,323.0078)(4,322.2786)(5,321.6146)(6,321.0885)(7,320.8646)(8,320.6432)(9,320.4271)(10,320.2604)(11,320.1406)};
\addlegendentry{control}
\end{axis}
\end{tikzpicture}
\caption{Anytime curves of the same-domain standard-UCT arm and the control, as pooled mean makespan over all 96 held-out cells---including the 24 zero-headroom dense cells, a wider pool than the $n=72$ conditional domain of Table~\ref{tab:uct-strata}---per budget, with $x$ giving $\log_2$ budget. Standard UCT descends with budget---a genuine search improvement, in contrast to the original rule's nearly flat curve---but the control descends slightly faster and edges ahead from about $B=8$ onward, ending marginally lower, that is better. The two curves finish within $0.2\%$ of each other, consistent with the non-rejection at the primary budget.}
\label{fig:uct-anytime}
\end{figure}
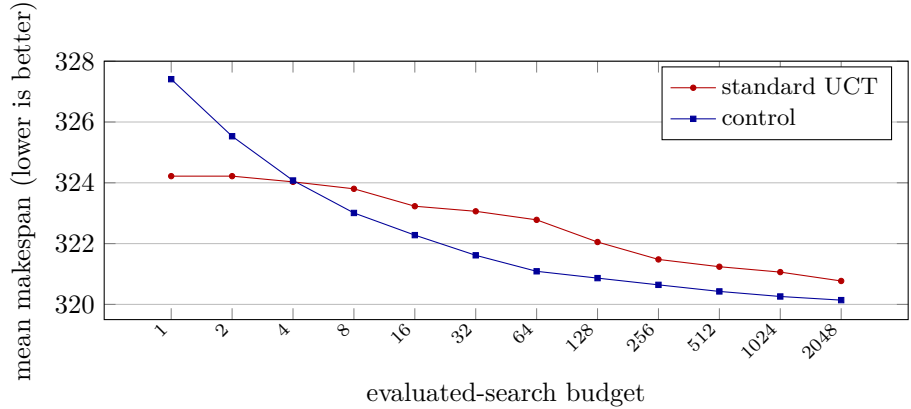

\textbf{Result.} At the primary budget $B=2048$ on the conditional domain, $H>\varepsilon$ with $n=72$, standard UCT and the randomized-restart control are not distinguishable at $\alpha=0.05$: the paired estimand $\Delta=(\mathrm{control}-\mathrm{uct})/\mathrm{control}$ has mean $-0.26\%$ with bootstrap 95\% CI $[-0.71\%,+0.23\%]$, Hodges--Lehmann $-0.24\%$, and UCT win/tie/loss $=14/25/33$, and no family stratum survives Holm correction (Table~\ref{tab:uct-strata}, Figure~\ref{fig:uct-forest}); note that the bootstrap intervals in Table~\ref{tab:uct-strata} are not Holm-corrected while the $p$-values are, so a stratum interval that excludes zero, as fork-join's does, need not be significant after correction. This is a bounded non-rejection rather than an equivalence result: we ran no equivalence test, and what the interval supports is that a UCT advantage beyond $+0.23\%$, or a disadvantage beyond $0.71\%$, lies outside the 95\% interval. The stratum directions are heterogeneous---fork-join $-0.71\%$, sparse-wide $-0.69\%$, layered $+0.63\%$; the dense family has $n=0$ under the headroom filter and, as in the main analysis, still enters the Holm family with $p=1$. Unlike the original rule, whose anytime curve is nearly flat, standard UCT \emph{does} improve with budget (Figure~\ref{fig:uct-anytime}); it simply does not improve enough to overtake independent restarts under an equal budget.

\textbf{Reading.} Textbook UCT selection does not rescue MCTS in this domain under an equal-budget protocol, and the comparison separates two questions that a single headline number conflates. Most of the \emph{magnitude} of the observed deficit is attributable to that selection rule: swapping in canonical UCT moves the pooled estimate from $-2.88\%$ to $-0.26\%$, which is consistent with reading the selection collapse as rule-specific. The \emph{direction} of the conclusion is not attributable to it: canonical UCT still does not beat equal-budget independent restarts, and the interval places any UCT advantage at no more than $+0.23\%$. We therefore claim neither that standard UCT is significantly worse than the control---the pooled interval spans zero and no stratum passes correction---nor that a single-cause ``broken rule'' account explains away the negative direction. The registered arm upgrades the ``Is this just a broken MCTS?'' discussion of Section~\ref{sec:limitations} from clean-room, indirect, to same-domain, direct, evidence, without altering any frozen result.

\begin{table}[t]
\centering
\caption{Same-domain standard-UCT arm vs.\ randomized-restart control (precedence-aware model, $B=2048$, conditional domain $H>\varepsilon$). $\Delta=(\mathrm{control}-\mathrm{uct})/\mathrm{control}$; W/T/L $=$ UCT win/tie/loss per instance; HL $=$ Hodges--Lehmann; CI $=$ instance-bootstrap percentile with $10^4$ draws; $p$ $=$ paired sign-flip permutation, Holm-corrected across the four family strata, the empty dense stratum entering with $p=1$ as in the main analysis.}
\label{tab:uct-strata}
\begin{tabular}{lrrrrrr}
\toprule
Stratum & $n$ & W/T/L & mean $\Delta$ & HL & bootstrap 95\% CI & Holm $p$ \\
\midrule
fork-join   & 24 & 2/11/11 & $-0.71\%$ & $-0.26\%$ & $[-1.358\%,-0.082\%]$ & $0.174$ \\
sparse-wide & 24 & 2/9/13  & $-0.69\%$ & $-0.94\%$ & $[-1.318\%,+0.020\%]$ & $0.201$ \\
layered     & 24 & 10/5/9  & $+0.63\%$ & $+0.11\%$ & $[-0.264\%,+1.603\%]$ & $0.406$ \\
dense       & 0  & ---     & ---       & ---       & ---                   & --- \\
pooled      & 72 & 14/25/33 & $-0.26\%$ & $-0.24\%$ & $[-0.706\%,+0.229\%]$ & --- \\
\bottomrule
\end{tabular}
\end{table}

\section{Symmetric warm-start control arm (registered follow-up, completed)}
\label{app:symmetric-warmstart}
Section~\ref{sec:limitations} named the asymmetric warm start as an uncontrolled difference between the arms---not the only one, since tree structure and the tree's freedom over processor placement also differ---and argued it was conservative for our conclusion, registering a symmetric-correction experiment as an append-only supplement. That experiment has since been run under a preregistered protocol. We report it here as \emph{append-only} evidence: it introduces no change to any frozen confirmation-stage number, criterion, or conclusion in the main text.

\textbf{Protocol.} The warm-start control is the frozen control arm with exactly one change: the production CPOP schedule is prepended as evaluated search $\#1$, after which the remaining $B-1$ evaluated searches are the same uniform-random restarts, minimum earliest-finish-time processor selection, and strict-improvement incumbent rule as the frozen control. This mirrors the main arm's accounting, under which a warm-start replay is counted against the budget rather than credited as new search, and on the exploration matrix $700$ of $720$ of the $B=1$ incumbents are that warm start. The CPOP schedule is the production baseline, byte-identical to the warm start already carried by the main arm and by the standard-UCT arm of Appendix~\ref{app:standard-uct}. Everything else---held-out instances, seeds, independent verifier, precedence-aware model, and the budget ladder---is held fixed. We report three paired comparisons on the same held-out cells at the primary budget: a reproduction check, comparing the main arm with the empty-start control, which must reproduce the frozen headline; the primary symmetric comparison, comparing the main arm with the warm-start control; and a secondary control-versus-control comparison, warm-start versus empty-start, that isolates what the warm start is worth at the primary budget.

\textbf{Reproduction.} Re-running the main arm and the empty-start control from scratch with the committed binaries and the frozen manifest reproduces the confirmation-stage headline. The conditional estimand is $-0.0288269$ against the frozen $-0.0288269$, an agreement within $5\times10^{-10}$, with the same $n=72$, the same win/tie/loss of $7/7/58$, and the same per-stratum means, fork-join $-4.36\%$, sparse-wide $-4.26\%$, layered $-0.02\%$; the permutation $p$-values agree to Monte-Carlo error, with layered $0.961$. This corroborates the frozen number; it does not change it.

\textbf{Result.} Under the symmetric protocol the deficit persists and, if anything, widens (Table~\ref{tab:sym-strata}; Figure~\ref{fig:sym-forest}). The primary estimand $\Delta_{\mathrm{sym}}$---the gap of the main arm relative to the warm-start control---has mean $-2.93\%$ with bootstrap 95\% CI $[-3.67\%,-2.19\%]$, Hodges--Lehmann $-2.79\%$, win/tie/loss $4/9/59$, and $n=72$, against $-2.88\%$ for the asymmetric baseline. Both structured strata remain significant after Holm correction and favor the control, with fork-join $-4.36\%$ and sparse-wide $-4.26\%$; the layered stratum is again not significant, at $-0.17\%$ with Holm $p=1$. The control-versus-control comparison explains why the two protocols nearly coincide: at the primary budget the CPOP warm start is worth a mean $+0.05\%$ to the control, with 95\% CI $[0\%,+0.12\%]$, and it is worth exactly nothing on $69$ of $72$ cells, because independent restarts have already found schedules at least as good as CPOP; the three cells where it helps at all are layered.

\begin{figure}[t]
\centering
\begin{tikzpicture}
\begin{axis}[
  width=0.74\linewidth, height=5cm,
  xlabel={evaluated-search budget}, ylabel={mean makespan (lower is better)},
  xtick={0,1,2,3,4,5,6,7,8,9,10,11},
  xticklabels={1,2,4,8,16,32,64,128,256,512,1024,2048},
  x tick label style={font=\scriptsize,rotate=45,anchor=east},
  ymin=236, ymax=249, ymajorgrids,
  legend style={at={(0.98,0.98)},anchor=north east,font=\small},
  legend cell align=left,
]
\addplot[color=red!70!black, mark=*, mark size=1pt] coordinates {
 (0,244.2361)(1,244.2361)(2,244.2361)(3,244.2361)(4,244.2083)(5,244.1250)(6,243.9444)(7,243.6806)(8,243.6250)(9,243.6250)(10,243.6111)(11,243.4861)};
\addlegendentry{MCTS main arm}
\addplot[color=blue!60!black, mark=square*, mark size=1pt] coordinates {
 (0,248.6215)(1,245.5278)(2,242.8889)(3,240.8785)(4,239.7604)(5,238.8646)(6,238.3090)(7,237.9653)(8,237.6910)(9,237.4236)(10,237.2222)(11,237.0174)};
\addlegendentry{control (empty start)}
\addplot[color=green!45!black, mark=triangle*, mark size=1.2pt] coordinates {
 (0,245.4167)(1,242.7882)(2,241.0069)(3,239.7361)(4,238.8438)(5,238.2535)(6,237.8889)(7,237.5972)(8,237.3854)(9,237.1944)(10,237.0486)(11,236.9028)};
\addlegendentry{control (CPOP warm start)}
\end{axis}
\end{tikzpicture}
\caption{Anytime curves of the main arm, the empty-start control, and the CPOP-warm-started control, as pooled mean makespan over the $n=72$ conditional domain per budget, with $x$ giving $\log_2$ budget. The main arm is nearly flat---its search barely improves on the warm start---while both controls descend with budget. The warm-start control starts below the empty-start control, reflecting the CPOP head start, but the two converge by the primary budget, and both finish well below the main arm. Symmetrizing the warm start does not change the endpoint ordering.}
\label{fig:sym-anytime}
\end{figure}
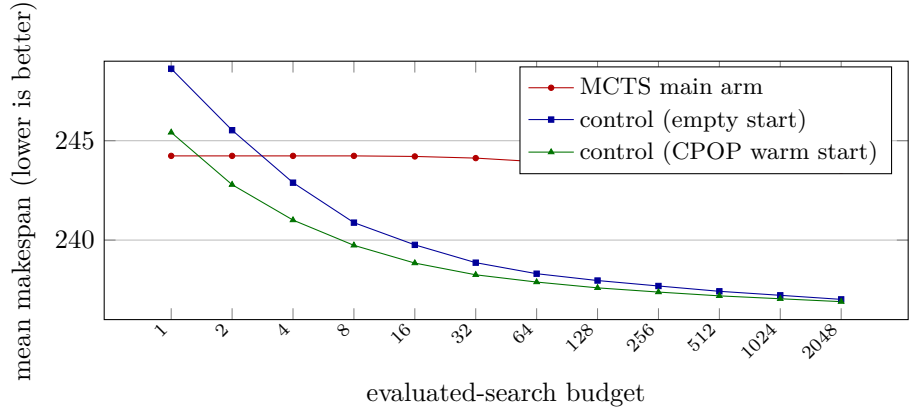

\textbf{Reading.} Removing the warm-start asymmetry confirms that it was conservative for the conclusion. Handing the control the same CPOP warm start does not rescue the main arm; it moves the deficit from $-2.88\%$ to $-2.93\%$. The anytime view (Figure~\ref{fig:sym-anytime}) makes the mechanism visible: the main arm's curve is nearly flat, its search barely improving on the warm start it was handed, whereas both control curves descend steeply with budget. The warm-start control starts below the empty-start control---the CPOP head start---but the two converge by the primary budget, and both finish well below the main arm. The deficit is therefore a property of what the tree search does with its budget rather than of which arm was seeded.

\begin{table}[t]
\centering
\caption{Symmetric warm-start comparison: main arm vs.\ the CPOP-warm-started control (precedence-aware model, $B=2048$, conditional domain $H>\varepsilon$). $\Delta_{\mathrm{sym}}=(\text{warm-start control}-\text{main arm})/(\text{warm-start control})$; W/T/L $=$ main-arm win/tie/loss per instance; HL $=$ Hodges--Lehmann; CI $=$ instance-bootstrap percentile with $10^4$ draws; $p$ $=$ paired sign-flip permutation with $10^4$ draws, Holm-corrected across the four family strata, the empty dense stratum entering with $p=1$ as in the main analysis. Negative favors the control.}
\label{tab:sym-strata}
\begin{tabular}{lrrrrrr}
\toprule
Stratum & $n$ & W/T/L & mean $\Delta_{\mathrm{sym}}$ & HL & bootstrap 95\% CI & Holm $p$ \\
\midrule
fork-join   & 24 & 0/0/24  & $-4.36\%$ & $-4.35\%$ & $[-5.35\%,-3.39\%]$ & $<10^{-3}$ \\
sparse-wide & 24 & 0/2/22  & $-4.26\%$ & $-4.02\%$ & $[-5.46\%,-3.14\%]$ & $<10^{-3}$ \\
layered     & 24 & 4/7/13  & $-0.17\%$ & $-0.21\%$ & $[-0.95\%,+0.70\%]$ & $1.0$ \\
dense       & 0  & ---     & ---       & ---       & ---                 & --- \\
pooled      & 72 & 4/9/59  & $-2.93\%$ & $-2.79\%$ & $[-3.67\%,-2.19\%]$ & --- \\
\bottomrule
\end{tabular}
\end{table}

\begin{figure}[t]
\centering
\begin{tikzpicture}
\begin{axis}[
  width=0.66\linewidth, height=4.4cm,
  xlabel={$\Delta_{\mathrm{sym}}$ (\%), mean with bootstrap 95\% CI},
  xmin=-6, xmax=1.5,
  ytick={1,2,3,4}, yticklabels={pooled,fork-join,sparse-wide,layered},
  ymin=0.5, ymax=4.5,
  xmajorgrids, tick align=inside, enlarge y limits=false,
]
\draw[dashed] (axis cs:0,0.5) -- (axis cs:0,4.5);
\addplot+[only marks, mark=*, black, mark size=1.6pt,
  error bars/.cd, x dir=both, x explicit]
 coordinates {
  (-2.9304,1) +- (0.7354,0.7398)
  (-4.3645,2) +- (0.9829,0.9701)
  (-4.2601,3) +- (1.2033,1.1219)
  (-0.1665,4) +- (0.7788,0.8674)
 };
\end{axis}
\end{tikzpicture}
\caption{Symmetric warm-start comparison: per-stratum mean $\Delta_{\mathrm{sym}}$ with bootstrap 95\% CI, and the pooled estimate ($B=2048$, $H>\varepsilon$). Negative favors the control. The fork-join and sparse-wide intervals lie entirely below zero and survive Holm correction; the layered interval crosses zero. The pooled estimate $-2.93\%$ is marginally more negative than the asymmetric baseline $-2.88\%$.}
\label{fig:sym-forest}
\end{figure}
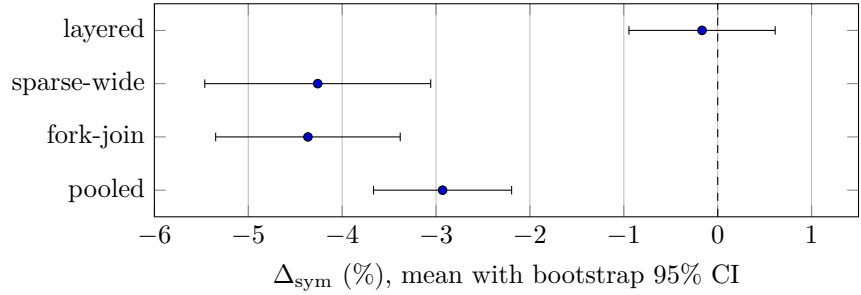

\section{Selection-distribution measurement (registered follow-up, completed)}
\label{app:selection}
Section~\ref{sec:limitations} registered a selection-artifact experiment---a direct measurement of the tree's selection distribution---as follow-up work, noting that until it was run the mechanism behind the deficit remained a hypothesis. That experiment has since been run under a preregistered protocol. We report it here as \emph{append-only} evidence: the selection distribution is orthogonal to makespan, so this section introduces no change to any frozen confirmation-stage number, criterion, or conclusion in the main text; it measures the selection distribution directly, while the causal step from that distribution to the makespan deficit remains an inference rather than a measured quantity.

\textbf{Protocol.} We instrument the main arm's selection step with a probe that, at fixed checkpoints, records the visit-count histogram over the root's children and derives two quantities: the normalized selection entropy $S_{\mathrm{root}}=-\sum_i p_i\log p_i/\log b$, where $p_i$ is child $i$'s visit share and $b$ is the branch width, so $S_{\mathrm{root}}=1$ is uniform, and the top-1 visit share. We run the probe on a frozen subset of the confirmation domain, $12$ instances with seeds $\{7,42\}$ and budgets $\{256,2048\}$, against two arms: the \emph{main} arm exactly as reported, and a \emph{value-repaired canary} that restores a finite value initialization. The subset and the read-out criteria were fixed before data collection. All $96$ runs completed with zero failures; raw per-run records and the entropy-trajectory figure are in the released bundle.

\textbf{Result.} The measurement reverses the direction the preregistration predicted, and we report it as registered. Far from spreading its trials, the reported rule \emph{concentrates} them: its late-stage normalized entropy is $0.29$ and its mean top-1 visit share is $0.86$, rising toward $1.0$ as the root fills, and it is near-uniform, with $S_{\mathrm{root}}>0.95$, on none of the $48$ late-stage cells, that is $0/48$. The value-repaired canary is the near-uniform arm instead, with $S_{\mathrm{root}}=1.00$ and top-1 $\approx 1/b$, near-uniform on $48/48$; the two arms' entropies separate by a paired mean of $+0.71$ over $48/48$ cells. Reading the visit histograms in child-creation order, the reported arm's most-visited child is the first-expanded one in $48/48$ late-stage runs---a deterministic first-child lock-in---whereas the canary's mass is scattered across creation indices. The concentration cross-checks against the original implementation's rank-$0$ selection signature: rank-$0$ share $0.80$ for the reported arm versus $0.28$ for the canary, consistent with the top-1 shares above (Table~\ref{tab:selection}).

\textbf{Mechanism.} The reported arm's value term is $\text{default\_Q}/Q_{v_i}$; a zero-initialized minimum makespan sets $Q_{v_i}=0$ on first evaluation, so the term diverges for every child regardless of $\text{default\_Q}$. Under strict-greater comparison, ties among diverging scores resolve to the first child in scan order, and once the root is full the fallback score for new children lets the first-created child keep winning---hence the first-child lock-in, a value-blind concentration rather than near-uniform exploration. Repairing the value channel does not yield a healthy selector either: the exploitation term is then of order $10$ while the $10^{5}$ exploration constant makes the exploration term of order $10^{4}$, so exploration dominates and the choice reverts to near-uniform. Both regimes are consistent with the nearly flat anytime curve and the $0.24\%$ self-improvement: added budget buys the tree almost nothing because it either barely explores or explores without exploiting.

\textbf{Reading.} The selection collapse behind the headline deficit is now measured, not conjectured; the causal step from that collapse to the makespan deficit itself remains an inference. Direct measurement shows the reported rule does not degenerate toward near-uniform choice; it collapses onto a single, value-blind branch, so its repeated trials are correlated and carry little new information. This revises the preregistered near-uniform conjecture into a measured first-child lock-in rather than leaving the earlier reading unchanged, and it leaves every frozen number untouched.

\begin{table}[t]
\centering
\caption{Selection-distribution measurement: the reported main arm vs.\ a value-repaired canary (frozen confirmation-domain subset, $96$ runs). $S_{\mathrm{root}}$: normalized selection entropy ($1$ $=$ uniform); top-1: largest child visit share; near-uniform: fraction of late-stage cells with $S_{\mathrm{root}}>0.95$; first-child: fraction whose most-visited child is the first-expanded one; rank-$0$: the original implementation's cross-check share.}
\label{tab:selection}
\footnotesize
\begin{tabular}{@{}l@{\hspace{1.5em}}rrrrr@{}}
\toprule
Arm & $S_{\mathrm{root}}$ (late) & top-1 (late) & near-uniform & first-child & rank-$0$ \\
\midrule
Reported main arm     & $0.29$ & $0.86$        & $0/48$  & $48/48$   & $0.80$ \\
Value-repaired canary & $1.00$ & $\approx 1/b$ & $48/48$ & scattered & $0.28$ \\
\bottomrule
\end{tabular}
\end{table}

\bibliographystyle{tmlr}
\bibliography{references}

\end{document}